\documentclass[10pt]{article}
\usepackage{fullpage}
\usepackage{spconf,amsmath,epsfig}
\usepackage{array,enumitem,setspace}
\usepackage[caption=false,font=footnotesize]{subfig}

\begin{document}\sloppy

\def\x{{\mathbf x}}
\def\L{{\cal L}}

\title{DISCO: Depth Inference from Stereo using Context}
%
\name{Kunal Swami, Kaushik Raghavan\sthanks{He was an intern from Indian Institute of Technology Madras}, Nikhilanj Pelluri\sthanks{He was an intern from BITS Pilani, Hyderabad}, Rituparna Sarkar, Pankaj Bajpai}
\address{Samsung Research Institute Bangalore, India}

\maketitle

\begin{abstract}
Recent deep learning based approaches have outperformed classical stereo matching methods. However, current deep learning based end-to-end stereo matching methods adopt a generic encoder-decoder style network with skip connections. To limit computational requirement, many networks perform excessive down sampling, which results in significant loss of useful low-level information. Additionally, many network designs do not exploit the rich multi-scale contextual information. In this work, we address these aforementioned problems by carefully designing the network architecture to preserve required spatial information throughout the network, while at the same time achieve large effective receptive field to extract multiscale contextual information. For the first time, we create a synthetic disparity dataset reflecting real life images captured using a smartphone; this enables us to obtain state-of-the-art results on common real life images. The proposed model DISCO is pre-trained on the synthetic Scene Flow dataset and evaluated on popular benchmarks and our in-house dataset of challenging real life images. The proposed model outperforms existing state-of-the-art methods in terms of quality as well as quantitative metrics.
\end{abstract}

\section{Introduction}
\label{sec:intro}
Depth information is extensively used in computer vision applications, such as autonomous vehicles, augmented reality, 3D reconstruction and simulation of artistic effects, such as bokeh effect. A cost effective depth estimation approach employs a stereo camera setup in which two cameras separated by a certain distance (baseline) capture the same scene. The captured images display a spatial shift in pixels along the baseline direction which is inversely proportional to the distance of the corresponding 3D world point from the camera plane. The correspondence matching step (main and non-trivial step) takes a rectified \cite{adv_comput_stereo_2003} stereo image pair and for each pixel $(x,y)$ in one view, identifies a matching pixel $(x\pm\delta,y)$ in the second view. Here, $\delta$ is called disparity of the pixel. The depth $z$ of a pixel $i$ is computed as follows:

\begin{equation}
z(i) = \frac{f * B}{\delta (i)}
\label{eq:intro}
\end{equation}

\newcommand{\samplewidth}{0.24}
\begin{figure}[t]
	\centering
	\subfloat[Input (face masked for anonymity)]{\includegraphics[width=\samplewidth\textwidth]{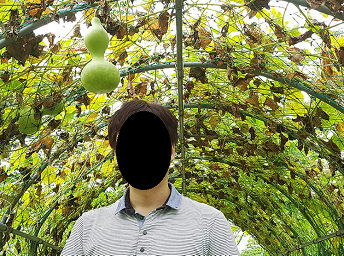}}
	\hfill
	\subfloat[DISCO result]{\includegraphics[width=\samplewidth\textwidth]{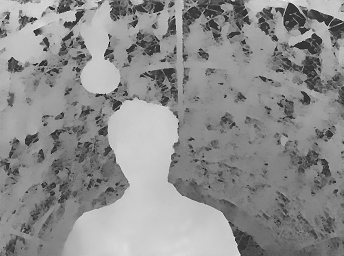}}
	
	\caption{Sample results of DISCO on a real image.}
	\label{fig:samplerealresult}
\end{figure}

Here, $f$ and $B$ denote the focal length and baseline of the stereo camera setup. 

A typical stereo correspondence matching task comprises of following four steps \cite{taxonomy_schar_ijcv_2002}: matching cost computation, cost aggregation, disparity calculation and disparity refinement. Accurate and robust correspondence matching is a challenging problem. Presence of occlusions, smooth regions, repetitive patterns, transparent objects, radiometric differences, such as illumination and color noise between left and right images and reflective and/or non-Lambertian surfaces pose major challenge \cite{adv_comput_stereo_2003}. Efficacy of classical stereo matching methods is limited by choice of hand-crafted features and filtering methods. Recently convolutional neural networks (CNN) have shown promising results in stereo depth estimation. In literature, the existing deep learning based stereo matching methods can be broadly classified into following categories:

\subsection{CNN based Cost Matching}
\label{subsec:cnn_cost_matching_methods}
These methods focused on improving the matching cost computation step \cite{taxonomy_schar_ijcv_2002}. It was first attempted by Zbontar and Lecun \cite{zbontar_jmlr_2016} who employed Siamese network architecture to compute matching cost between two image patches. \cite{urtasun_cvpr_2016} improved the computational complexity of \cite{zbontar_jmlr_2016}, whereas \cite{highway_networks_shaked_cvpr_2017} additionally incorporated confidence estimation in their network. These methods are limited, as they are mainly used to compute matching cost, while disparity estimation and refinement is still performed using classical methods.

\subsection{CNN based End-to-End Methods}
\label{subsec:cnn_end_to_end_methods}
Mayer \textit{et al.} \cite{dispnet_mayer_cvpr_2016} first proposed an end-to-end disparity estimation network called DispNet. They also offered a large synthetic dataset called Scene Flow with dense disparity ground-truth. \cite{crl_iccvw_2017} and \cite{iresnet_cvpr_2018} improved DispNet by adding a second stage disparity correction network. \cite{edgestereo_2018} integrated an explicit edge detection network to improve disparity estimation, while \cite{segstereo_eccv_2018} proposed a unified segmentation and disparity estimation architecture in order to incorporate semantic information. Kendal \textit{et al.} \cite{gcnet_iccv_2017} were first to use 3D convolutions for end-to-end disparity estimation, their network GC-Net estimates a disparity cost volume and a differentiable soft argmin function is used to regress the final disparity map.  \cite{explicit_cost_aggr_aaai_2018} and \cite{psmnet_cvpr_2018} extended GC-Net by adding an explicit cost aggregation subnetwork and spatial pyramid pooling respectively. However, 3D convolutions are computationally very demanding \cite{3d_convs_tpami_2013} which is reflected in the runtime of these methods.

Most of the current end-to-end disparity estimation networks are generic encoder-decoder style networks; the encoder part computes abstract features and decoder part progressively up samples and estimates disparity map at each of these up sampled resolution. The encoder part uses a high down sampling rate ($1/64$\textsuperscript{th} of input resolution) to extract high-level contextual information, while a randomly cropped image patch $256$ (H) x $512$ (W) (limited by the hardware) is used for training. We argue that this kind of design has inherent limitations for dense per-pixel disparity estimation task:

\begin{enumerate}[label=(\alph*)]
	\itemsep0em
	\item Excessive down sampling by encoder results in loss of spatial information crucial for disparity estimation.
	\item Smaller effective receptive field of the network, combined with low spatial resolution fails to capture required dense contextual information.
\end{enumerate}

In this work, we address these limitations, we present \emph{DISCO---\textbf{D}epth \textbf{I}nference from \textbf{S}tereo using \textbf{Co}ntext}, a novel end-to-end disparity estimation model in which every encoder block extracts dense multiscale contextual information. Additionally, the overall down sampling rate is restricted to $1/16$\textsuperscript{th} of original resolution. To the best of our knowledge, this is the first work in literature which makes use of dense blocks and dilated convolutions to achieve this kind of novel design for an end-to-end disparity estimation network. Furthermore, current public datasets \cite{dispnet_mayer_cvpr_2016, kitti_2015, eth3d_2017} contain stereo images with very large disparity range. Hence, models trained on these datasets fail to achieve desirable results on real life images captured by a smartphone. Our synthetic dataset, which is first of its kind in literature, enables us to achieve state-of-the-art results on these real images (see Fig.~\ref{fig:samplerealresult}). Following are the major contributions of this work:
\begin{itemize}
	\itemsep0em
	\item We use dense blocks as the basic building block of the disparity estimation network to effectively preserve low-level spatial information throughout the network. Also, we use different increasing dilation rates in the dense blocks so that each encoder block effectively captures dense contextual information at different receptive fields\textemdash as large as 128.
	\item We propose a novel \emph{local and global context fusion} (LGCF) module for cost volume computation.
	\item We conduct extensive experiments and ablation studies to demonstrate advantages our network design choices.
	\item Extensive evaluation and comparison of DISCO on challenging benchmark datasets \cite{dispnet_mayer_cvpr_2016,middlebury_2014,eth3d_2017} show that it significantly outperforms recent state-of-the-art.
	\item For the first time, we create a synthetic dataset reflecting real life scenarios captured by a smartphone.
\end{itemize}

\newcommand{\netSize}{0.90}
\begin{figure*}[t]
	\centering
	\subfloat{\includegraphics[width=\netSize\textwidth]{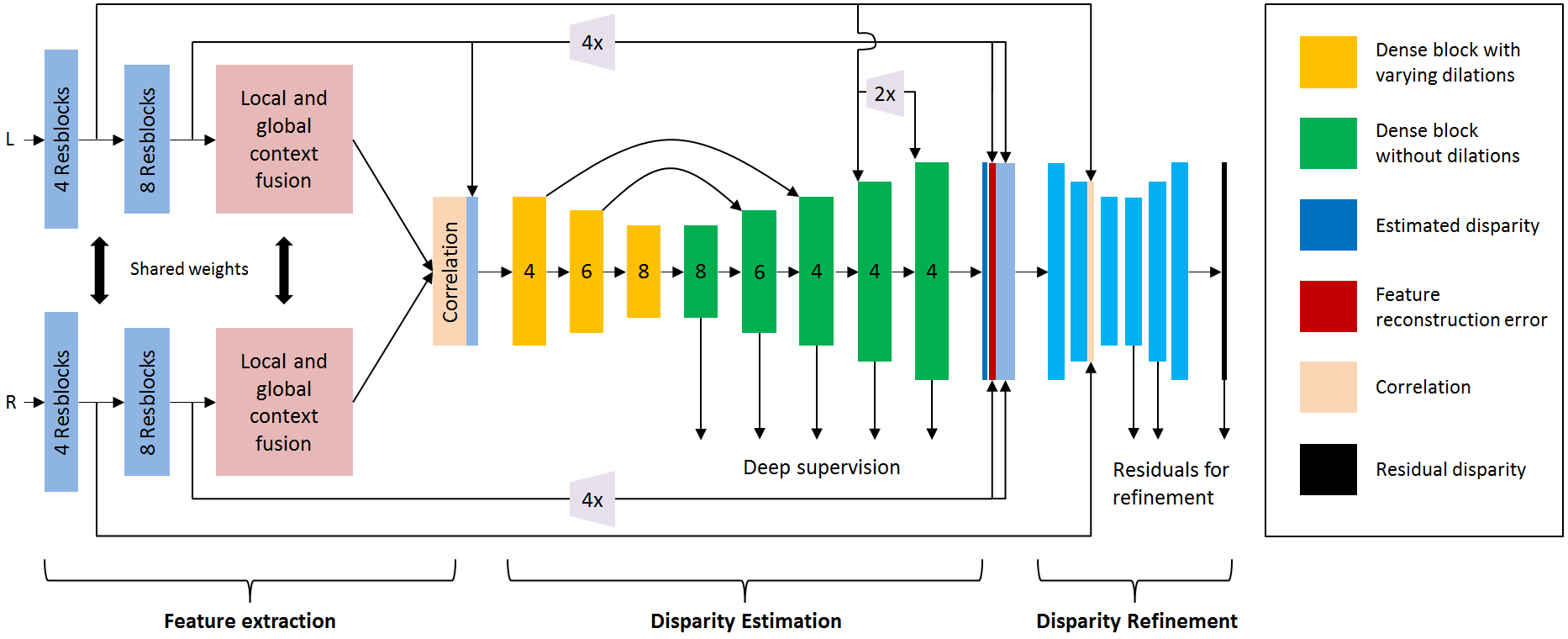}}
	
	\caption{Architecture of the proposed model DISCO (best viewed in color).}
	\label{fig:modelarchitecture}
\end{figure*}

\section{Model Architecture}
\label{sec:model_arch}
We first present an overview of DISCO, then explain its important modules and our design choices.

\subsection{Architecture Details}
\label{subsec:arch_details}
Fig.~\ref{fig:modelarchitecture} shows the detailed architecture of the proposed model, it can be conceptually divided into three subnetworks:

\subsubsection{Feature Extraction}
\label{subsubsec:feature_ext}
The feature extraction module is a Siamese network architecture comprising of $12$ residual blocks, first $4$ blocks operate at ${1/2}^{th}$ original resolution, while subsequent 8 blocks operate at ${1/4}^{th}$ resolution. The last block of feature extraction module is a dense local and global feature fusion module (see Section~\ref{subsec:lgcf}) aimed at improving the cost volume computation. The cost volume is created with maximum disparity range of $80$ pixels which corresponds to a displacement of $320$ at full resolution and is chosen based on the disparity range of public stereo datasets \cite{dispnet_mayer_cvpr_2016, kitti_2015, middlebury_2014, eth3d_2017}. The cost volume is concatenated with left image feature maps and forms the input of the disparity estimation subnetwork.

\subsubsection{Disparity Estimation}
\label{subsubsec:disp_est}
This is an encoder decoder network with dense block as the basic building block. Each of the three encoder blocks down samples its input by a factor of $2$. Thus, maximum down sampling rate is ${1/16}^{th}$; this preserves the low-level spatial information effectively than existing methods where maximum down sampling rate is $1/64$. Dense blocks in the encoder part contain convolution layers with different dilation rates; thus, unlike conventional encoder-decoder architecture, each encoder block in our network captures dense multiscale contextual information (see Section~\ref{subsec:dilated_convs}). Five decoder blocks estimate disparity map at ${1/16}^{th}$, ${1/8}^{th}$, ${1/4}^{th}$, half and full resolution; this enables deep supervision\textemdash improved gradient flow to initial network layers. Skip connections are added between corresponding encoder and decoder blocks; the last two decoder blocks receive multiscale low-level feature information from feature extraction module.

\subsubsection{Disparity Refinement}
\label{subsubsec:disp_ref}
This is a generic encoder-decoder network similar to \cite{iresnet_cvpr_2018} aimed at refining the estimated disparity map. It has three convolution and deconvolution layers. The input to this subnetwork is a concatenation of the estimated disparity map, feature reconstruction error and up sampled left image feature maps. The feature reconstruction error (or warping error) \cite{unsupervised_garg_eccv_2016} imposes an extra geometric constraint. The estimated left image disparity map is used to generate a warped left image from right image and the difference between the warped and original left image constitutes the warp error. We use high-level left and right image feature maps instead of original input images to achieve increased robustness. Finally, the estimated residuals by refinement subnetwork are added to the corresponding disparity maps from estimation network.

\subsection{Dense Block}
\label{subsec:dense_block}
In a traditional CNN, the output $x_l$ of $l^{th}$ convolution layer is obtained by applying a non-linear transformation function $H_l$ to the output of previous $(l-1)^{th}$ convolution layer. In DenseNet \cite{densenet_cvpr_2017}, the output of $l^{th}$ convolution layer is obtained by applying $H_l$ to concatenation of feature maps from all previous layers. This kind of dense connectivity ensures implicit deep supervision, improves feature propagation and encourages feature reuse. Hence, DenseNet is an ideal choice for the dense per-pixel disparity estimation task. 

DenseNet comprises of several dense blocks, where each dense block contains $l$ layers. The non-linear transformation $H_{l}$ of each layer is a composite function of: batch-normalization, rectified linear unit (ReLU) and a 3x3 convolution. Each layer of a dense block produces $g$ feature maps; hence, the output of $l^{th}$ layer of a dense block has $g_{0}+(l-1)*g$ feature maps where $g_{0}$ is the number of input feature maps of the dense block. The parameter $g$, called growth rate, keeps the number of parameters of the network in check. 

In our design, $H_{l}$ is a composite function of only two operations: an exponential linear unit (ELU) \cite{elu_resnet_vjti_2016} and a 3x3 convolution. The ELU activation function is defined as follows:

\begin{equation}
elu(x) = 
\begin{cases}
x,               & x < 1\\
\alpha(e^{x}-1), & \text{otherwise}
\end{cases}
\label{eq:elu}
\end{equation}

\subsection{Dilated Convolution and Effective Receptive Field}
\label{subsec:dilated_convs}
Dilated convolutions are a good trade-off to increase the effective receptive field without increasing the number of network parameters. A convolution kernel of size $k$ x $k$ with dilation rate $d$ achieves an effective receptive field $rC_{k, d} = (k-1)(d-1) + k$. We capitalize on our network design choice of dense blocks and set different increasing dilation rates in each encoder block of our disparity estimation subnetwork. As an example, an encoder block with 4 convolution layers and dilation rates of 1, 3, 6 and 8 respectively captures information at 4 varying receptive fields\textemdash 3, 7, 13 and 17; the effective receptive field of this encoder block is:

\begin{equation}
rC_{3, 1} + rC_{3, 3} + rC_{3, 6} + rC_{3, 8} - 3 = 37
\label{eq:effective_receptive_field}
\end{equation}

Thus, each encoder block in our network captures dense contextual information at different scales. A similar module was recently introduced in \cite{denseaspp} for semantic segmentation.

\begin{figure}[t]
	\centering
	\subfloat{\includegraphics[width=1.0\linewidth]{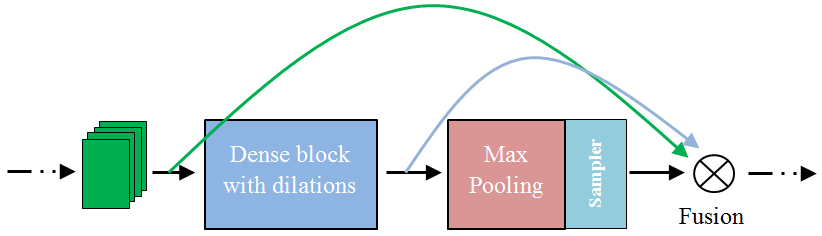}}
	
	\caption{Local and Global Context Fusion (LGCF) module.}
	\label{fig:lgcf}
\end{figure}

\subsection{Local and Global Context Fusion for Correlation}
\label{subsec:lgcf}
The correlation layer operation is conceptually similar to normalized cross-correlation \cite{adv_comput_stereo_2003} in traditional methods. However, the correlation layer \cite{dispnet_mayer_cvpr_2016} simply performs multiplicative patch comparisons between the left and right feature maps at each disparity level. In existing methods, this correlation operation is simply applied with a kernel size of 1x1. Thus, it does not consider even the neighborhood contextual information and is vulnerable to noise. Na\"{\i}vely increasing the kernel size has the limitation that the receptive fields remain fixed.

In this work, we propose a deep local and global context fusion module (see Fig.~\ref{fig:lgcf}). This module consists of a dense block with 6 convolution layers and a spatial pyramid pooling (SPP) block with 4 max-pooling layers. The dense block submodule is aimed at extracting long range spatial information, while the SPP submodule is aimed at extracting the global contextual information at four different scales. Each convolution layer in the dense block has dilation rate of 1, 3, 6, 12, 18 and 24 respectively; thus, achieving an effective receptive field of 126 which covers full height and half width of the training input image at $1/4^{th}$ resolution. The max-pooling layers in the SPP block have square kernel size of 8, 16, 32 and 64 respectively; thus, effectively capturing the global contextual information at four varying scales. The local and global contextual information from these two submodules are fused along with original deep features using 1x1 convolutions before performing the correlation operation.	

\subsection{Loss Function}
\label{subsec:loss_func}
DISCO is trained with Huber loss function defined as follows:
\begin{equation}
L(D,D_{gt}) = 
\begin{cases}
0.5t^2, &  \parallel t \parallel_1 < 1\\
\parallel t \parallel_1-0.5,              & \text{otherwise}
\end{cases}
\label{eq:smoothl1}
\end{equation}

Here, $D$ and $D_{gt}$ are estimated and ground-truth disparities respectively and $t = D - D_{gt}$.

\section{Experimental Setup}
\subsection{Public Datasets}
\label{subsec:datasets}

Scene Flow \cite{dispnet_mayer_cvpr_2016} is a large synthetic dataset consisting of 35454 training images and 4370 test images. Our model is pre-trained on Scene Flow and later fine-tuned on other datasets to prevent over-fitting (other datasets have less training images).

Middlebury \cite{middlebury_2014} dataset consists of real world indoor images. Both training and testing sets have 15 images each. The disparity range of this dataset is $\approx$400 pixels (for half resolution images).

ETH3D \cite{eth3d_2017} dataset contains a variety of challenging real world images. The training set consists of 27 images, while the test set has 20 images. The disparity range is $\approx$80 pixels.

KITTI 2015 \cite{kitti_2015} has challenging outdoor driving scenes. Both training and testing sets consist of 200 images each and the disparity range is $\approx$240 pixels.

\begin{figure}[t]
	\subfloat{
		\includegraphics[width=0.23\linewidth]{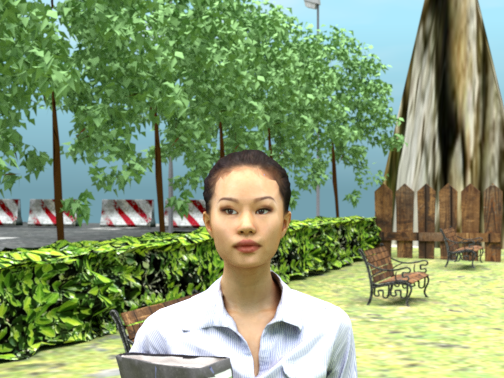}	
		\includegraphics[width=0.23\linewidth]{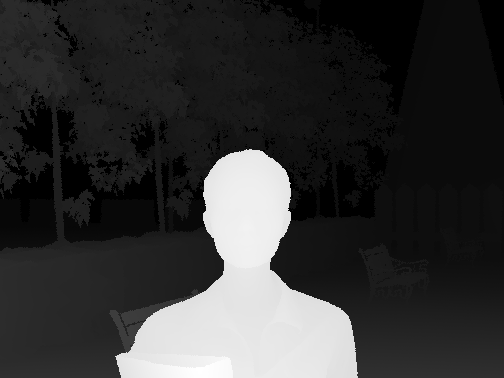}
	}
	\hspace{0.01\linewidth}
	\subfloat{
		\includegraphics[width=0.23\linewidth]{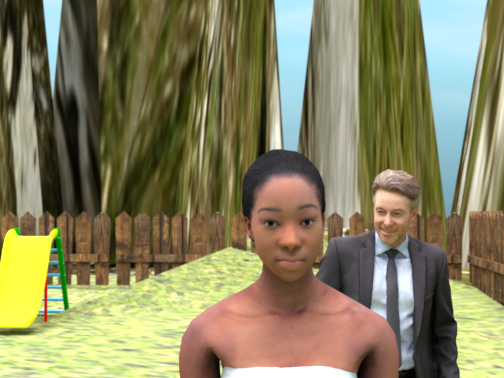}
		\includegraphics[width=0.23\linewidth]{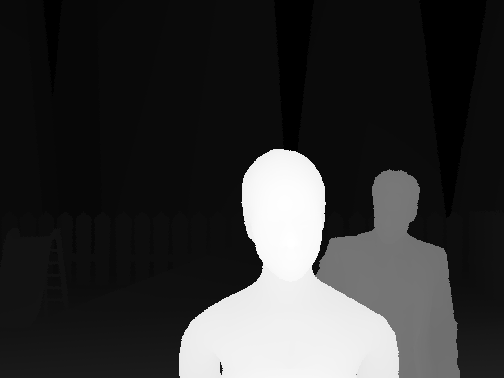}
	}
	\caption{Representative images from our synthetic dataset}
	\label{fig:ssd_representative}	
\end{figure}

\subsection{Our Dataset}
\label{subsubsec:ssd}
In this work, we created our own synthetic dataset of $\approx$10000 stereo images with dense disparity and depth ground-truth. The dataset consists of high quality realistic scenes reflecting real life smartphone capture scenarios.  Fig.~\ref{fig:ssd_representative} shows some representative images. We use the open source software Blender \cite{blender} to create our dataset, we omit the detailed discussion of our dataset creation because of page constraint.

\begin{figure*}
	\subfloat[Input]{\includegraphics[width=0.16\linewidth]{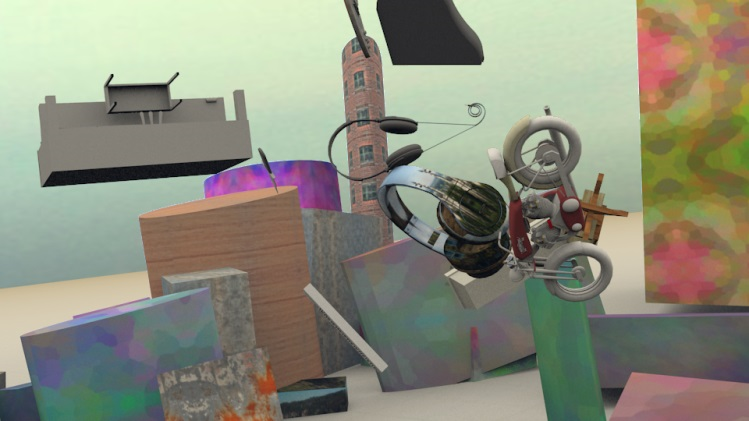}}	
	\hspace{0.001\linewidth}
	\subfloat[Ground-truth]{\includegraphics[width=0.16\linewidth]{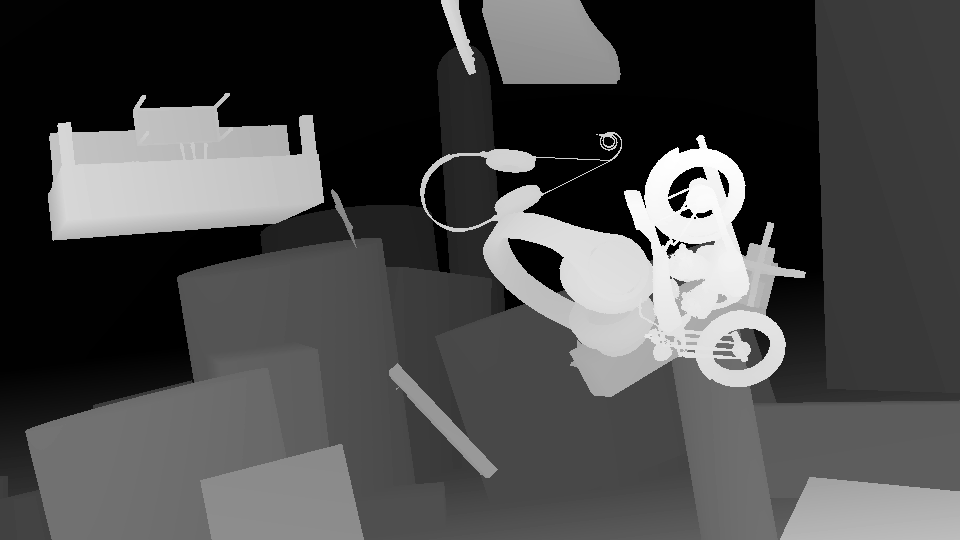}}
	\hspace{0.001\linewidth}
	\subfloat[DISCO]{\includegraphics[width=0.16\linewidth]{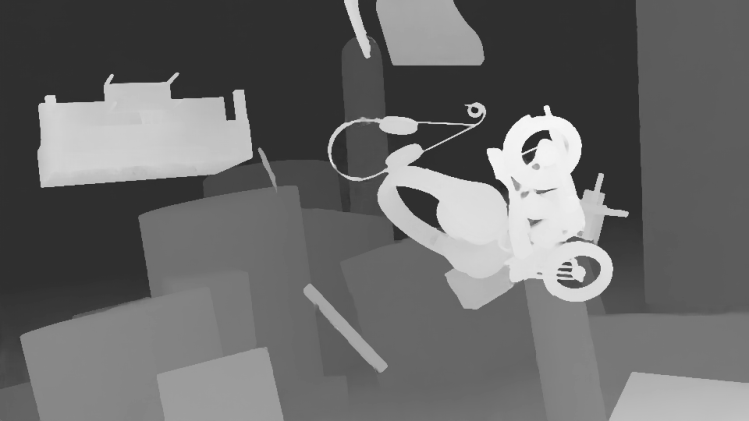}}	
	\hspace{0.001\linewidth}
	\subfloat[B+D+C]{\includegraphics[width=0.16\linewidth]{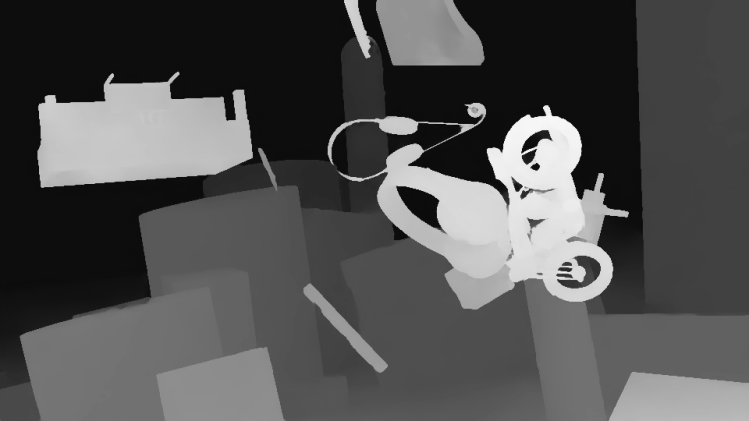}}
	\hspace{0.001\linewidth}
	\subfloat[Baseline+Dilations]{\includegraphics[width=0.16\linewidth]{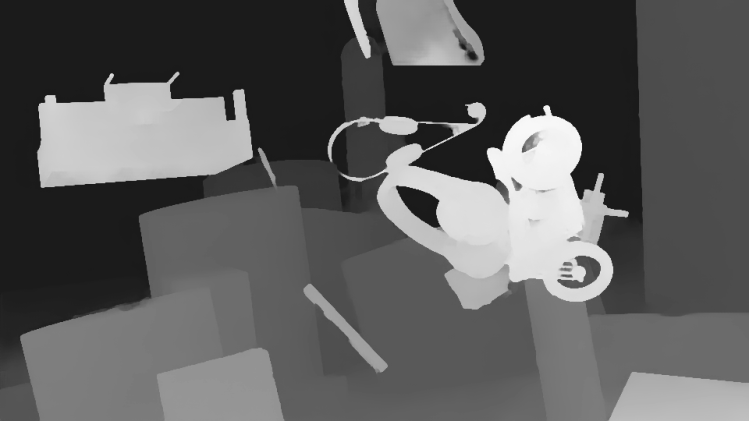}}	
	\hspace{0.001\linewidth}
	\subfloat[Baseline]{\includegraphics[width=0.16\linewidth]{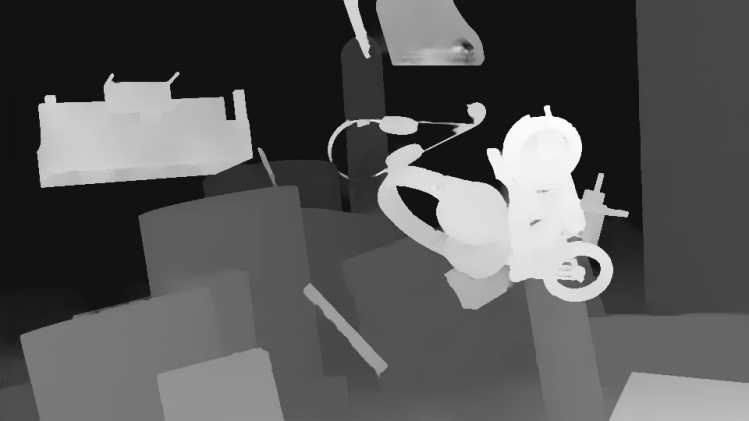}}
	\caption{Qualitative comparison between models variants in our ablation study on Scene Flow test dataset. B+D+C stands for Baseline+Dilations+Context}
	\label{fig:ablation_sceneflow}	
\end{figure*}

\subsection{Training Details}
\label{subsec:training_details}
We use TensorFlow framework, Adam optimization was used with initial learning rate 2x10-4 and momentum 0.9. Stepwise learning rate decay was applied. Our model was pre-trained on Scene Flow dataset till 200 epochs using batch-size 12 on 4 NVIDIA P40 GPUs, image size for training was 256 (H) x 512 (W). Data augmentation in the form of random crop, brightness, gamma and color shift was applied on the fly.

\subsection{Evaluation Metrics}
\label{subsec:metrics}
End-Point Error (EPE) is the mean absolute difference for all pixels between the estimated and ground-truth disparity maps; whereas 3-Pixel Error (3PE) denotes the percentage of estimated disparity pixels whose absolute difference from ground-truth disparity is greater than 3.

\section{Results and Discussions}
\label{sec:results}
First, we show the effectiveness of our network design choices by reporting results of ablation studies. We evaluate and compare following model variants: \textbf{Baseline}: This model doesn't contain any of the proposed module\textemdash dilated convolutions in encoder dense blocks, LGCF and the refinement subnetwork. \textbf{Baseline + Dilations}: Increasing dilation rates are introduced in convolution layers of encoder dense blocks. This is to justify the benefit of multiscale contextual information in the encoder. \textbf{Baseline + Dilations + Context}: We additionally introduce the LGCF module to demonstrate its benefit in cost volume computation. \textbf{DISCO:} This model additionally includes the refinement subnetwork and is the final proposed model. Table~\ref{table:ablation_studies} compares these four model variants, while Fig.~\ref{fig:ablation_sceneflow} shows qualitative comparison. It can be observed that adding increasing dilation rates in the encoder dense blocks significantly improves the results ($\approx$6\%), whereas LGCF module alone improves both the error metrics by at least 20\%. Thus, dense multiscale contextual information is useful in the computation of cost volume which forms the main input of disparity estimation subnetwork. Lastly, the refinement subnetwork slightly improves the results.

\begin{table}[t]
	\caption{Quantitative comparison between different model variants on Scene Flow test dataset}
	\begin{tabular}{|m{0.75cm} | m{1.2cm} | m{1.6cm} | m{1.65cm} | m{1.2cm} | }
		\hline
		& \textbf{Baseline} & 
		\textbf{Baseline + \newline Dilations} &
		\textbf{Baseline + \newline Dilations + \newline Context} & 
		\textbf{DISCO} \\
		\hline
		\textit{EPE} & \hfil 1.23  & \hfil 1.16 & \hfil 0.96 & \hfil \textbf{0.91} \\
		\hline
		\textit{3PE} & \hfil 5.32  & \hfil 4.91 & \hfil 4.01 & \hfil \textbf{3.85} \\
		\hline
	\end{tabular}\\
	\label{table:ablation_studies}
\end{table}

\begin{table}[t]
	\caption{Quantitative comparison of DISCO with existing state-of-the-art methods on KITTI 2015 test dataset}
	\begin{tabular}{|m{0.5cm} | m{1.0cm} | m{0.85cm} | m{1.5cm} | m{1.2cm} | m{1cm} |}
		\hline
		& 
		\textbf{DispNet \cite{dispnet_mayer_cvpr_2016}} & 
		\textbf{LRCR \cite{lrcr_cvpr_2018}} &
		\textbf{EdgeStereo \cite{edgestereo_2018}} & 
		\textbf{GC-Net \cite{gcnet_iccv_2017}} & 
		\textbf{DISCO} \\
		\hline
		\textit{3PE} & \hfil 4.34  & \hfil 3.03 & \hfil 2.99 & \hfil 2.87 & \hfil \textbf{2.86} \\
		\hline
	\end{tabular}\\
	\label{table:kitti_results}
\end{table}

\begin{table}[t]
	\caption{ETH3D leaderboard top-5 (Oct. 14, 2018)}
	\begin{tabular}{|m{0.5cm} | m{1.3cm} | m{0.9cm} | m{1.2cm} | m{1.0cm} | m{1.1cm} |}
		\hline
		& 
		\textbf{DN-CSS} & 
		\textbf{LALA} &
		\textbf{iResNet} & 
		\textbf{DLCB} & 
		\textbf{DISCO} \\
		\hline
		\textit{99P} & \hfil 2.89  & \hfil 2.88 & \hfil 2.72 & \hfil 2.55 & \hfil \textbf{2.40} \\
		\hline
	\end{tabular}\\
	\label{table:eth_results}
\end{table}

\begin{table}[t]
	\caption{Middlebury leaderboard top-5 (Oct. 14, 2018)}
	\begin{tabular}{|m{0.5cm} | m{1.2cm} | m{0.9cm} | m{1.1cm} | m{1.0cm} | m{1.3cm} |}
		\hline
		& 
		\textbf{PSMNet} & 
		\textbf{NOSS} &
		\textbf{iResNet} & 
		\textbf{DISCO} & 
		\textbf{DN-CSS} \\
		\hline
		\textit{99P} & \hfil 106  & \hfil 104 & \hfil 87.5 & \hfil 86.6 & \hfil \textbf{82.0} \\
		\hline
	\end{tabular}\\
	\label{table:middlebury_results}
\end{table}

\begin{table*}[t]
	\caption{Quantitative comparison of DISCO with existing state-of-the-art methods on Scene Flow test dataset; *PSMNet considers disparity range of 192, so for fair comparison we additionally report our EPE of 0.76 following their criteria}
	\newcolumntype{"}{@{\hskip\tabcolsep\vrule width 2pt\hskip\tabcolsep}}\makeatother
	\begin{tabular}{|m{1.4cm}|m{1.2cm}|m{1.2cm}|m{1.4cm}|m{1.6cm}|m{1.4cm}|m{1.25cm}|m{1.25cm}|m{1.4cm}|m{1.25cm}|}
		\hline
		& 
		\textbf{DispNet\linebreak \cite{dispnet_mayer_cvpr_2016}} & 
		\textbf{CRL\linebreak \cite{crl_iccvw_2017}} & 
		\textbf{iResNet\linebreak \cite{iresnet_cvpr_2018}} & 
		\textbf{EdgeStereo\linebreak \cite{edgestereo_2018}} & 
		\textbf{SegStereo\linebreak \cite{segstereo_eccv_2018}} & 
		\textbf{GC-Net\linebreak \cite{gcnet_iccv_2017}} &
		\textbf{Yu et al.\linebreak \cite{gcnet_iccv_2017}} &
		\textbf{PSMNet\linebreak \cite{psmnet_cvpr_2018}} &
		\textbf{DISCO} \\
		\hline
		\textit{EPE} & \hfil 1.68 & \hfil 1.32 & \hfil 1.40 & \hfil 1.11 & \hfil 1.45 & \hfil 2.51 & \hfil 1.75 & \hfil 1.09* & \textbf{0.91/0.76*} \\
		\hline
		\textit{3PE} & \hfil - & \hfil 6.20 & \hfil 4.93 & \hfil 4.97 & \hfil - & \hfil 9.34 & \hfil 5.62 & \hfil - & \hfil \textbf{3.85} \\
		\hline
	\end{tabular}\\
	\label{table:sceneflow_test}
\end{table*}

Finally, we evaluate DISCO against existing state-of-the-art methods. Table~\ref{table:kitti_results} quantitatively compares state-of-the-art methods from KITTI 2015 leaderboard, DISCO outperforms four methods, viz., DispNet \cite{dispnet_mayer_cvpr_2016}, LRCR \cite{lrcr_cvpr_2018}, EdgeStereo \cite{edgestereo_2018} and GC-Net \cite{gcnet_iccv_2017}. Table~\ref{table:sceneflow_test} shows quantitative comparison of DISCO with recent state-of-the-art methods on Scene Flow test dataset. It can be observed that DISCO achieves EPE and 3PE measures which are significantly lower than other methods. Since PSMNet \cite{psmnet_cvpr_2018} only considers disparities $\leq$ 192 for evaluation, for fairness we additionally report EPE of 0.76 following their procedure. In summary, our EPE is more than 22\% and 3PE is more than 28\% better (lower) than the next best method in Table~\ref{table:sceneflow_test}.

\begin{figure*}[t]
	\centering
	\subfloat{\includegraphics[width=0.15\linewidth]{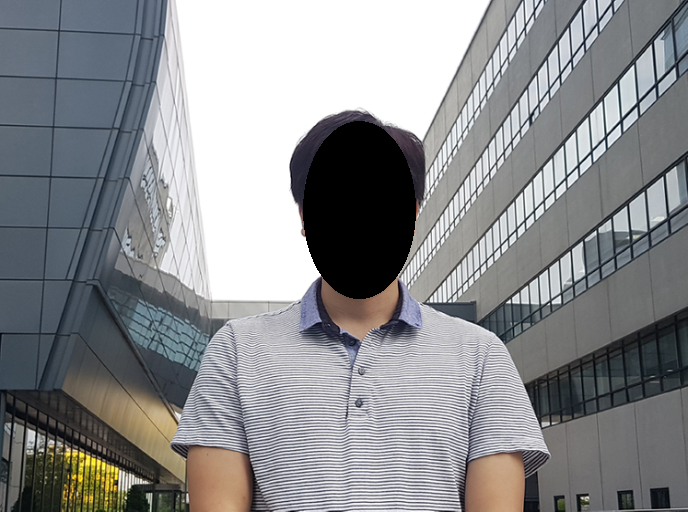}}
	\hfil
	\subfloat{\includegraphics[width=0.15\linewidth]{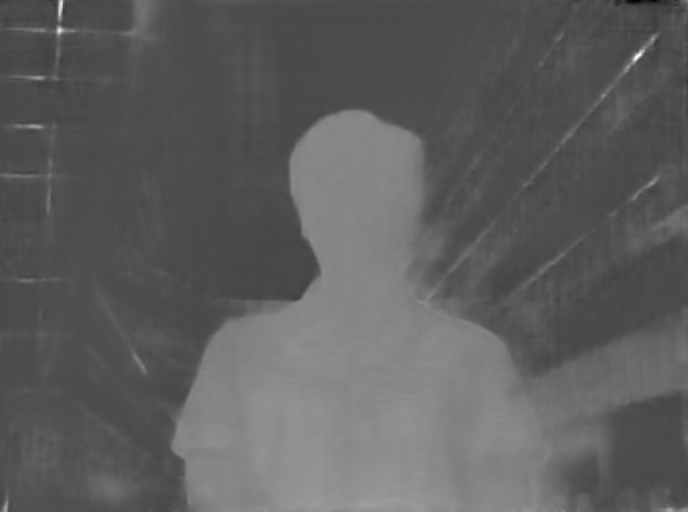}}
	\hfil
	\subfloat{\includegraphics[width=0.15\linewidth]{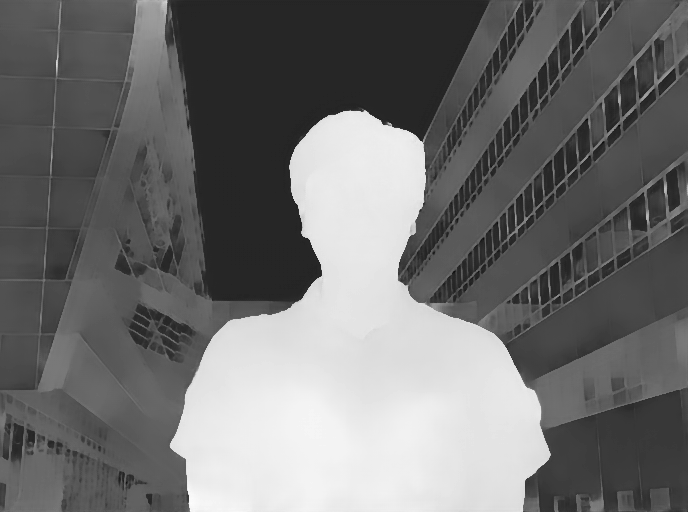}}
	\hfil
	\subfloat{\includegraphics[width=0.15\linewidth]{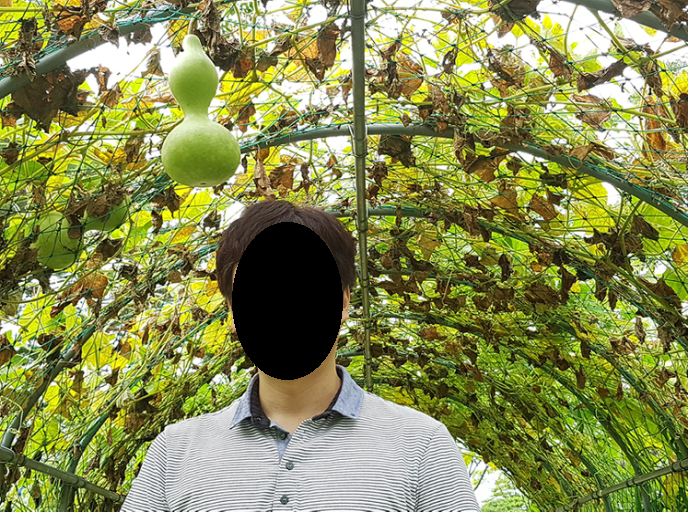}}
	\hfil
	\subfloat{\includegraphics[width=0.15\linewidth]{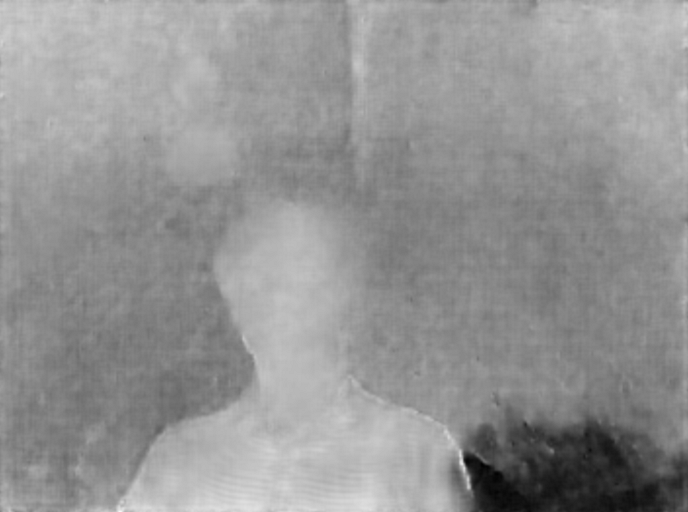}}
	\hfil
	\subfloat{\includegraphics[width=0.15\linewidth]{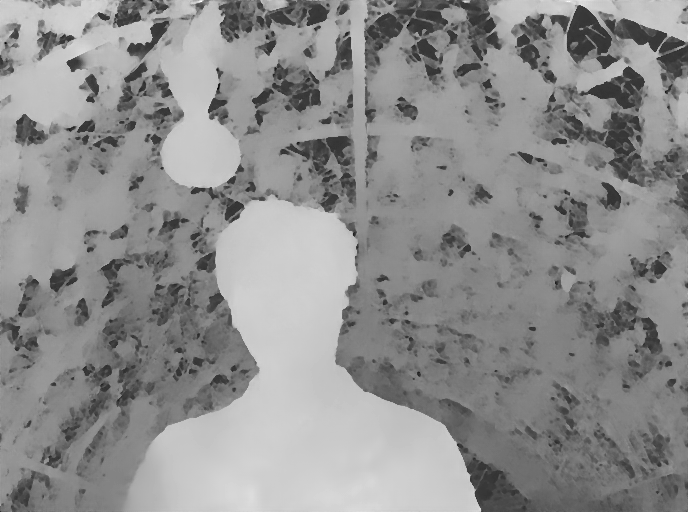}}
	
	\vfil
	
	\setcounter{subfigure}{0}
	\subfloat[Input]{\includegraphics[width=0.15\linewidth]{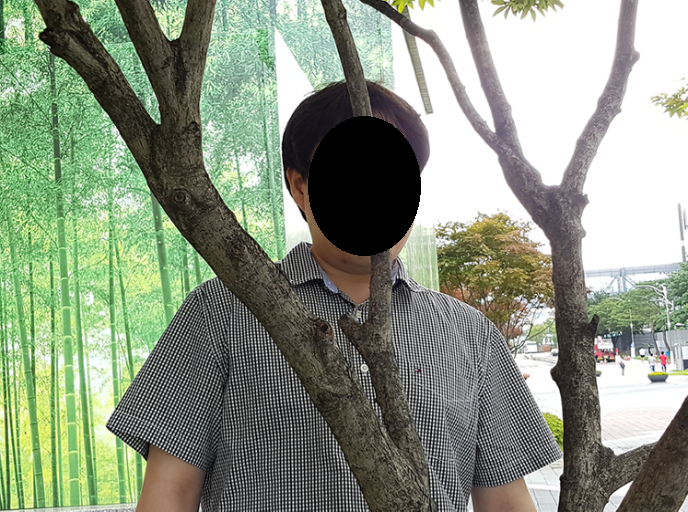}}
	\hfil
	\subfloat[DispNet \cite{dispnet_mayer_cvpr_2016}]{\includegraphics[width=0.15\linewidth]{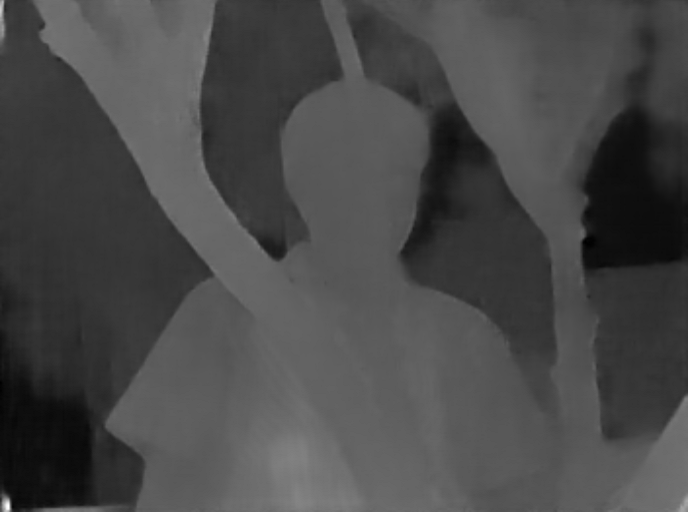}}
	\hfil
	\subfloat[DISCO]{\includegraphics[width=0.15\linewidth]{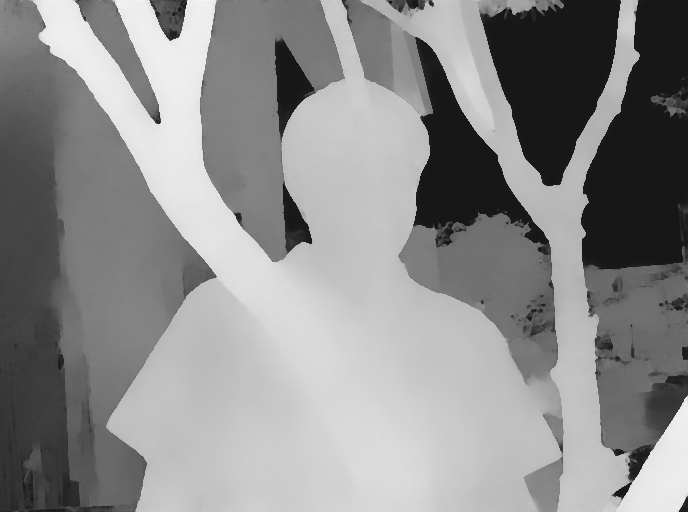}}
	\hfil
	\subfloat[Input]{\includegraphics[width=0.15\linewidth]{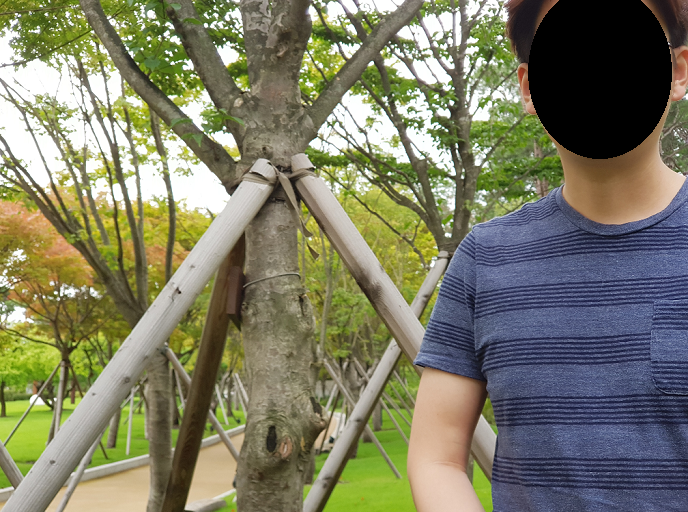}}
	\hfil
	\subfloat[DispNet \cite{dispnet_mayer_cvpr_2016}]{\includegraphics[width=0.15\linewidth]{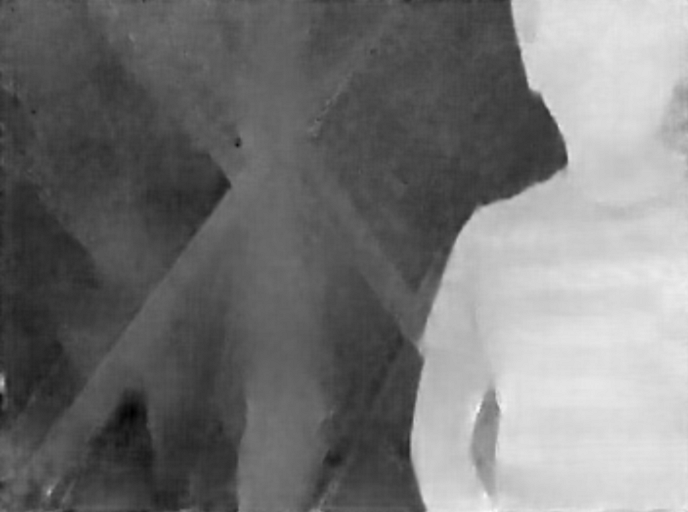}}
	\hfil
	\subfloat[DISCO]{\includegraphics[width=0.15\linewidth]{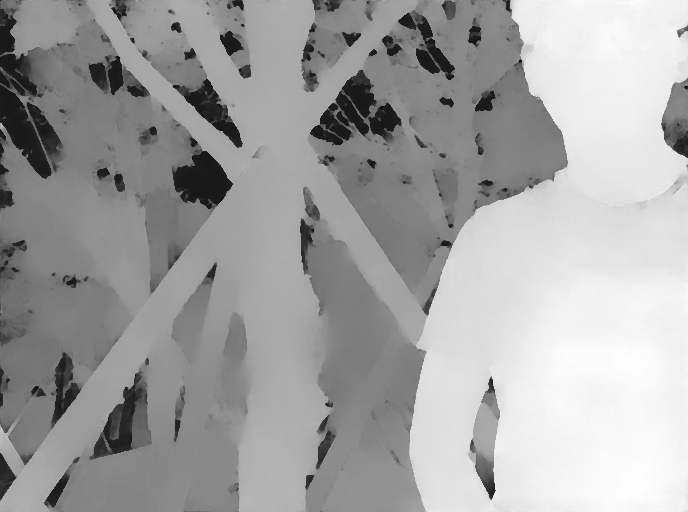}}
	
	\caption{Results of DISCO on some of our challenging real images (no ground-truth). Faces are masked to preserve anonymity.}
	\label{fig:real_results}
\end{figure*}

Table~\ref{table:eth_results} and Table~\ref{table:middlebury_results} show the top-5 methods on ETH3D and Middlebury official leaderboard for a given metric. DISCO tops the ETH3D leaderboard, while it achieves second rank in Middlebury. Fig.~\ref{fig:real_results} shows results of DISCO on challenging real life images captured by a smartphone; existing methods, such as DispNet \cite{dispnet_mayer_cvpr_2016} fail to obtain desirable results on these real life images.

\section{Conclusions}
\label{sec:conclusion}
A novel end-to-end deep learning based disparity estimation model was proposed. The proposed model was uniquely designed to extract long range spatial and contextual information at every part of the network. A novel local and global context fusion module was designed to effectively compute stereo cost volume. Systematic experiments were performed and the advantage of each proposed module was demonstrated. Additionally, a synthetic dataset reflecting real life smartphone captures was generated for the first time. The proposed model significantly outperformed existing state-of-the-art on both synthetic and real benchmarks. The proposed model also achieved state-of-the art results on real life smartphone captured images on which most existing models fail.

\bibliographystyle{IEEEbib}
\small{
\bibliography{stereo_refs}
}

\end{document}